\documentclass[journal,twoside,web]{ieeecolor}
\usepackage{generic}
\usepackage{cite}
\usepackage{amsmath,amssymb,amsfonts}
\usepackage{algorithmic}
\usepackage{graphicx}
\usepackage{algorithm,algorithmic}
\usepackage{hyperref}
\usepackage{multicol}
\usepackage{multirow}
 \usepackage{booktabs}

\usepackage{tikz}
\usepackage{subcaption}

\usepackage{tabularx}
\usepackage{multicol}
\usepackage{caption}
\usepackage{array}
\usepackage{lmodern}
\usepackage{booktabs}

\usetikzlibrary{positioning, chains, shapes.geometric, fit, shapes, arrows.meta, calc}

\hypersetup{hidelinks=true}
\usepackage{textcomp}
\def\BibTeX{{\rm B\kern-.05em{\sc i\kern-.025em b}\kern-.08em
    T\kern-.1667em\lower.7ex\hbox{E}\kern-.125emX}}
\begin{document}
\title{\textsc{AttentionDep}: Domain-Aware Attention for Explainable Depression Severity Assessment}

\author{Yusif Ibrahimov,Tarique Anwar, Tommy Yuan, Turan Mutallimov and Elgun Hasanov
\thanks{Yusif Ibrahimov is with 
the Department of Computer Science, University of York, Heslington, United Kingdom (e-mail:yusif.ibrahimov@york.ac.uk) and French-Azerbaijani University under Azerbaijan State Oil and Industry University (e-mail:yusif.ibrahimov@ufaz.az)}
\thanks{Tarique Anwar is with 
the School of Computing Technologies, RMIT University, Melbourne, Australia (e-mail: tarique.anwar@rmit.edu.au).}
\thanks{Tommy Yuan is with 
the Department of Computer Science, University of York, Heslington, United Kingdom (e-mail: tommy.yuan@york.ac.uk).}
\thanks{Turan Mutallimov is with 
the Department of Engineering, University of Aberdeen, Scotland, United Kingdom (e-mail: t.mutallimov.22@abdn.ac.uk).}
\thanks{Elgun Hasanov is with 
the École Polytechnique, Institut Polytechnique de Paris, Palaiseau, France (e-mail: elgun.hasanov@polytechnique.edu).}
}
\maketitle

\begin{abstract}
In today's interconnected society, social media platforms provide a window into individuals' thoughts, emotions, and mental states. This paper explores the use of platforms like Facebook, $\mathbb{X}$ (formerly Twitter), and Reddit for depression severity detection. We propose \textsc{AttentionDep}, a domain-aware attention model that drives explainable depression severity estimation by fusing contextual and domain knowledge. Posts are encoded hierarchically using unigrams and bigrams, with attention mechanisms highlighting clinically relevant tokens. Domain knowledge from a curated mental health knowledge graph is incorporated through a cross-attention mechanism, enriching the contextual features. Finally, depression severity is predicted using an ordinal regression framework that respects the clinical-relevance and natural ordering of severity levels. Our experiments demonstrate that \textsc{AttentionDep} outperforms state-of-the-art baselines by over 5\% in graded $\text{F}_1$ score across datasets, while providing interpretable insights into its predictions. This work advances the development of trustworthy and transparent AI systems for mental health assessment from social media.

\end{abstract}


\section{Introduction}
\label{sec:introduction}

Depression affects over 280 million people globally, with severe outcomes, including approximately 700,000 suicides annually \cite{who2024depression}. Despite available treatments, barriers such as stigma and limited access of healthcare treatments leave over 70\% of affected individuals untreated \cite{olfson2016treatment}. The COVID-19 pandemic has intensified this crisis, highlighting the urgent need for effective and scalable methods for early symptom identification \cite{shader2020covid19}. Social media platforms such as Facebook, $\mathbb{X}$, and Reddit provide a rich source of user-generated content reflecting mental states, offering opportunities for automated depression assessment \cite{ghosh2021depression, anwar2022tracking}. Traditional interview- and questionnaire-based approaches, while informative, are resource-intensive and may lack scalability \cite{park2012depressive,park2013perception}.  

Recent research has explored social media-based models for depression detection \cite{ibrahimov2024explainable,yusif2025depressionx}. For example, Shen et al. \cite{shen2017depression} analysed a labelled $\mathbb{X}$ dataset for multimodal depression detection, while Sampath and Durairaj \cite{sampath2022data} employed Reddit data for severity-based classification using machine learning. Advanced deep learning approaches, such as the attention-based model by Naseem et al. \cite{naseem2022early} and DepressionNet \cite{zogan2021depressionnet}, which integrates text summarisation, further improve detection performance.  

Despite progress, two major gaps remain. First, most studies formulate depression detection as a binary classification problem \cite{shen2017depression,sadeque2018measuring,zogan2021depressionnet}, overlooking clinically relevant severity levels \cite{ibrahimov2024explainable}. Some recent models attempt regression or multi-class classification, but they often lack integration of domain-specific knowledge, limiting their clinical interpretability \cite{naseem2022early}. Second, the opacity of deep learning models complicates their adoption in healthcare settings, where transparency and explainability are essential for building trust among clinicians and users \cite{ibrahimov2024explainable}.  

To address these gaps, we propose \textsc{AttentionDep}, a domain-aware attention model for explainable depression severity detection from social media posts. \textsc{AttentionDep} follows a structured, multi-step approach. First, contextual representation learning encodes posts hierarchically using unigrams and bigrams. Attention mechanisms highlight clinically salient tokens, while cross-attention integrates domain knowledge from a Wikipedia-derived knowledge graph to enhance feature relevance. The knowledge graph captures clinical relations and provides embeddings that enrich the contextual representations. Finally, in the depression severity estimation stage, the infused representation is used to predict severity levels through an ordinal regression framework. It respects the inherent ordering of severity and improves clinical relevance. \textsc{AttentionDep} is inherently explainable, allowing insight into the decision-making process and providing transparency in predictions. Overall, we make the following contributions in this paper.

\begin{enumerate}
    \item We introduce \textsc{AttentionDep}, a domain-aware attention model that integrates contextual text features with domain knowledge for explainable depression severity prediction.  
    \item We develop a knowledge graph representation framework that captures post-specific clinical relations from Wikipedia, enhancing the semantic depth of input features.  
    \item We conduct extensive experiments and analyses to demonstrate the model’s effectiveness in accurately identifying depression severity levels.
    
    \item We provide explainability mechanisms to interpret model predictions, ensuring transparency and trustworthiness in clinical contexts.  
\end{enumerate}

The remainder of this paper is structured as follows. Section \ref{sec:relatedwork} covers related work, followed by the problem statement in Section \ref{sec:problemstatement}. Section \ref{sec:proposedmodel} presents the proposed model \textsc{AttentionDep}, and experimental results are presented in Section \ref{sec:experimentalresults}. Finally, Section \ref{sec:conclusion} concludes the paper.


\section{Related Work} \label{sec:relatedwork}

\noindent\textbf{\textit{Depression detection from social media}}. Social media platforms are valuable resources for detecting mental disorders such as depression \cite{zogan2021depressionnet,naseem2022early}, eating disorders \cite{abuhassan2023ednet,abuhassan2023classification} and suicide ideation \cite{naseem2022hybrid,sawhney2021towards}. eRisk shared task \cite{losada2018overview,losada2017erisk} by the Conference and Labs of the Evaluation Forum (CLEF)  focused on automatic depression detection in user posts. Other studies have analysed depression indicators through emotions \cite{park2012depressive,park2013perception}, linguistic style \cite{trotzek2018utilizing}, social interactions \cite{mihov2022mentalnet,pirayesh2021mentalspot}, and online behaviours \cite{de2013predicting,shen2017depression}. For instance, Park et al. \cite{park2012depressive} used interviews to link $\mathbb{X}$ users' language with depressive moods, while Shen et al. \cite{shen2017depression}  developed a labelled dataset from $\mathbb{X}$ for multimodal analysis. Deep learning (DL) models, including CNNs 
  \cite{jacovi2018understanding}, LSTMs 
  \cite{hochreiter1997long}, and GRUs \cite{Chung2014-ss}, have become effective tools in this domain. Trotzek et al. \cite{trotzek2018utilizing} explored CNNs with word embeddings and linguistic style, while 
  Zogan et al. \cite{zogan2021depressionnet} achieved high performance using automatic text summarisation. However, most approaches are limited to binary classification, ignoring severity levels. Naseem et al. \cite{naseem2022early} recently introduced a multi-class model with TextGCN and BiLSTM for severity detection. Graph neural networks (GNNs) have also been applied in mental disorder detection, capturing textual and social relationships \cite{naseem2022early,Liu2021-uk,pirayesh2021mentalspot}. Mihov et al. \cite{mihov2022mentalnet} proposed MentalNet, a GCN-based model utilizing heterogeneous graphs, while Pirayesh et al. 
  \cite{pirayesh2021mentalspot} introduced MentalSpot, based on friends' social posts and interactions.

\noindent\textbf{\textit{Explainability and depression detection}}. While DL models are effective, their complexity often limits transparency—a critical factor in healthcare. Explainable models like decision trees are self-explanatory, but DL models require post-hoc explainability techniques to clarify decisions \cite{Arya2019-yg,Jia2022-hp}. In depression detection, explainability is often underexplored, despite the disorder's complexity. Attention-based methods, including self-attention \cite{Amini2020-zk}, hierarchical attention \cite{han-etal-2022-hierarchical}, and multi-head attention models \cite{naseem2022hybrid}, have been used to address this need, providing insights into the model's reasoning. Moreover, DepressionX \cite{ibrahimov2025depressionx} is a recent explainable deep learning model for depression severity estimation that employs a knowledge-infusion from an external knowledge graph. However, it employs a common large knowledge graph for all the instances, and the resulting explainabilities are not very informative, as the words are too short and the sentences too abstract. Furthermore, Dalal et al. \cite{dalal2024cross} developed the PSAT (Process knowledge-infused cross Attention) framework for the explainable analysis of depression diagnostics by estimating PHQ-9–based depression symptoms, which yielded promising results. However, their model lacks generality as it heavily relies on depression-specific feature extraction, making it applicable only to depression analysis. Moreover, in order to capture all possible PHQ-9–based symptoms, PSAT requires detailed data from the user, which makes it impractical for final diagnostics of depression (existence or severity) when only limited posts are available.


\section{Problem statement} \label{sec:problemstatement}


The objective of this research is to estimate the severity of depression from social media posts in accordance with established clinical standards, specifically the Depressive Disorder Annotation (DDA) scheme~\cite{mowery-etal-2015-towards} and Beck’s Depression Inventory (BDI)~\cite{beck_ii}. Depression severity is defined as an ordered spectrum of four distinct levels: $C = \{minima, mild, moderate, severe\}$. Let $P = \{p_1, p_2, \ldots, p_{N}\}$ denote a set of social media posts. Each post $p_i \in P$, authored by a user $u_i \in U$, is associated with a ground-truth severity label $y_i \in C$. The research problem is thus formulated as learning a function
\[
f: P \rightarrow C, \quad f(p_i) = y_i,
\]
that automatically classifies each post $p_i$ into one of the four ordered severity levels.

\section{Proposed model} \label{sec:proposedmodel}

\begin{figure*}[!ht]
\begin{center}
 \includegraphics[width=\textwidth]{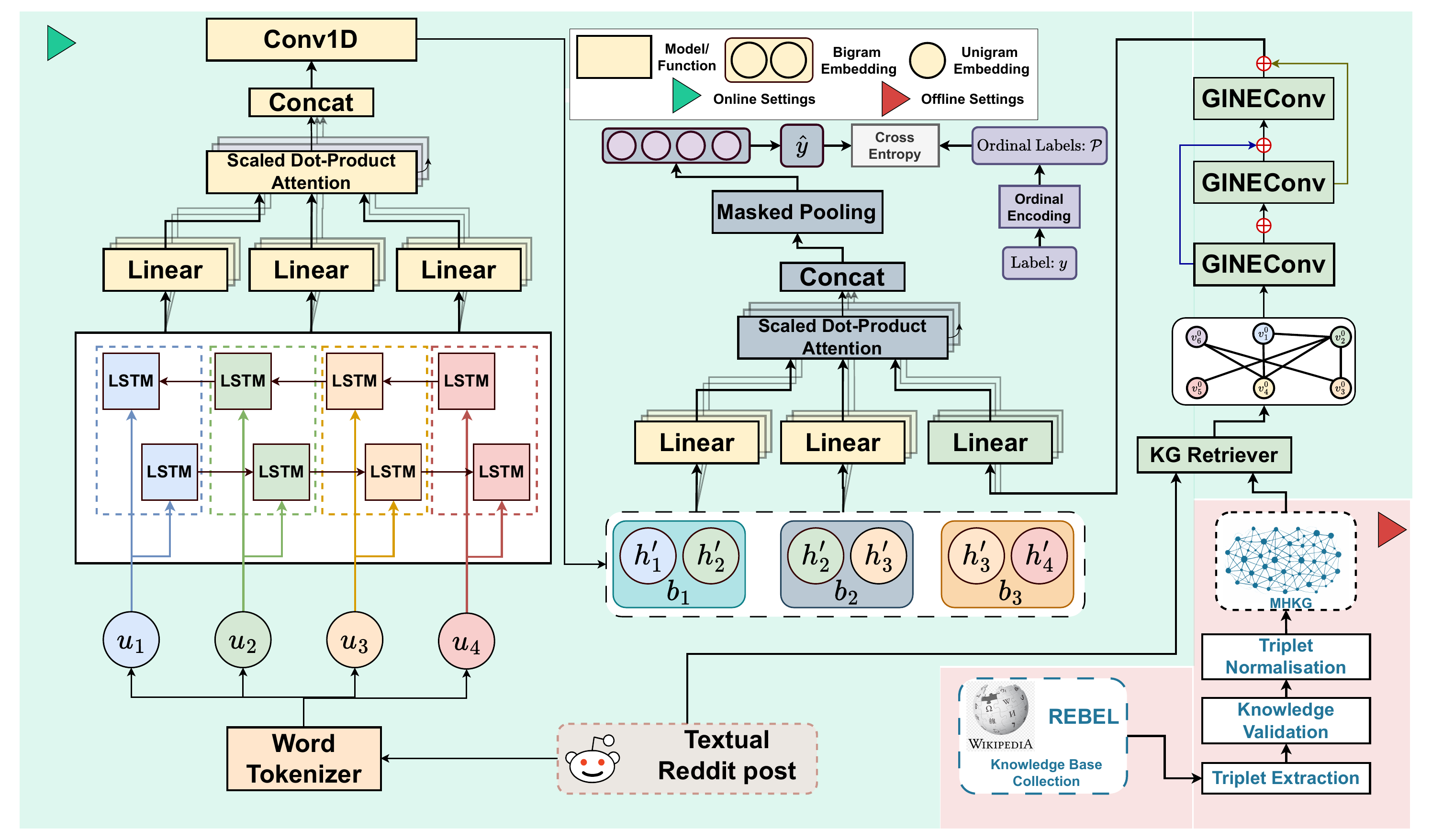}
  \caption{Proposed model \textsc{AttentionDep}. It predicts depression severity of a post and generates explanation with unigram- and bigram-level attentions and a knowledge graph.}
 \label{fig:depressionx}
\end{center}
\end{figure*}

We propose \textsc{AttentionDep}, a domain-aware attention model for explainable prediction of depression severity from social media posts. The model is designed in three main stages. First, user posts are encoded using unigram and bigram representations, with attention mechanisms to highlight clinically relevant tokens. Second, domain knowledge from a curated depression-specific knowledge graph is incorporated to enrich these representations. Cross-attention then integrates contextual and domain-informed features to create a knowledge-aware post representation. Finally, depression severity is predicted through an ordinal regression framework, respecting the natural ordering of severity levels (minimal, mild, moderate, severe). This architecture not only improves predictive accuracy but also provides interpretable insights into which textual and domain-specific features influenced each prediction. The overall model design is illustrated in Figure~\ref{fig:depressionx}.




\subsection{Context and domain knowledge modelling}

AttentionDep models textual representations of social media posts in a way that captures both the semantic meaning and depression-related expressions, while incorporating clinically relevant domain knowledge. The modelling process is hierarchical. It begins with unigram-level encoding, progresses through contextualisation and attention mechanisms that highlight linguistically and clinically salient tokens, and culminates in bigram-level representations enhanced via knowledge graph infusion. Together, these components enable the model to capture both the linguistic context and symptom-related signals of depression present in user posts.

\subsubsection{Contextual unigram encoding}
As the first step, social media posts are tokenised into unigrams using the NLTK library\footnote{\url{https://www.nltk.org/api/nltk.tokenize.html}}. For unigram-level encoding, we adopt FastText \cite{bojanowski-etal-2017-enriching}, a computationally efficient embedding model that integrates character-level n-grams with CBOW and skip-gram training. FastText represents words by composing subword information, making it effective for handling out-of-vocabulary (OOV) and misspelled tokens:  $\mathbf{u_w} \leftarrow \frac{1}{|G_w|}\sum_{g \in G_w} \mathbf{z_g} $, 
%
%
where $\mathbf{u_w}$ is the embedding for token $w$, $G_w$ is the set of its subword n-grams, and $\mathbf{z_g}$ is the embedding of subword $g$. The ability to model OOV words is particularly critical in our setting, since social media posts about depression often contain informal language, creative spellings, or deliberately obfuscated terms (e.g., to bypass moderation or stigma). We use 300-dimensional FastText embeddings pre-trained on 2 million words with subword information from the Common Crawl corpus (600B tokens)\footnote{\url{https://fasttext.cc/docs/en/english-vectors.html}}. Hence, for each post $p_i$, token embeddings are denoted $\mathbf{u}_{i,j} = \text{FastText}(w_{i,j}) \in \mathbb{R}^{300}$ for $j = 1, \dots, L$ tokens.

Understanding context is particularly crucial in depression-related text, since the same word can signal very different meanings depending on the author’s mental state and situational framing. For example, \textit{``I'm \textbf{tired} from a productive lecture''} expresses ordinary fatigue, whereas \textit{``I'm so \textbf{tired} of living''} conveys potential signs of hopelessness and emotional exhaustion. Although FastText generates a single static embedding for each word, it does not account for such contextual variation. To capture these nuances, we further encode the sequence of tokens using a bidirectional long short-term memory (BiLSTM) network, a recurrent neural architecture that models the context of each token based on its surrounding tokens in both forward and backward directions. The contextual unigram encodings are thus defined in Equation \ref{eqn:unigram}, where $d$ is the embedding dimension.  
\begin{eqnarray}
  \mathbf{H}_i \leftarrow \{\mathbf{h}_{i,j}\}_{j=1}^L = \text{BiLSTM}(\{\mathbf{u}_{i,j}\}_{j=1}^L) \in \mathbb{R}^{L \times d} \label{eqn:unigram}
\end{eqnarray}

\subsubsection{Attention to clinically salient unigrams}

In social media posts, not all words contribute equally to understanding a user’s mental health state. Certain tokens (e.g., \textit{``worthless''}, \textit{``empty''}, or \textit{``tired''}) often appear in posts indicative of moderate or severe depression, while many others (e.g., stopwords or neutral terms) carry little diagnostic value. To capture this variability, we incorporate an attention mechanism that learns to assign higher weights to words that are more predictive of depression severity labels during training \cite{galassi2020attention}. In this way, the model automatically learns which words are most informative for distinguishing between \textit{minimal}, \textit{mild}, \textit{moderate}, and \textit{severe} classes, without requiring explicit manual annotation of clinical keywords. We adopt a multi-head attention mechanism \cite{vaswani2017attention}, which computes weighted representations over the sequence of contextualised unigram embeddings. Each attention head captures different patterns of association between tokens and depression severity, thereby improving robustness and interpretability. The attention scores for head $k$ are computed as:
\begin{align}
    \mathbf{A}_i^k &\leftarrow \sigma \left( \frac{(\mathbf{H}_{i}\mathbf{W}^k_{\mathbf{Q}}) (\mathbf{H}_{i}\mathbf{W}^k_{\mathbf{K}})^T}{\sqrt{d_k}} \right) (\mathbf{H}_{i}\mathbf{W}^k_{\mathbf{V}}) \label{eqn:attn-head} \\
    \mathbf{H}'_i &\leftarrow \bigoplus_{k=1}^{\kappa_u} \left[ \mathbf{A}_i^k \right] \mathbf{W}_O \label{eqn:mha}
\end{align}
where $\mathbf{W}^k_{\mathbf{Q}}$, $\mathbf{W}^k_{\mathbf{K}}$, and $\mathbf{W}^k_{\mathbf{V}}$ are the query, key, and value projection matrices for head $k$, $\kappa_u$ is the number of attention heads, $\sigma$ is the softmax function, and $\mathbf{W}_O$ is a trainable projection matrix. The resulting representation $\mathbf{H}'_i$ captures unigram-level signals that are most predictive of depression severity. Importantly, the learned attention weights provide interpretable insights into which words the model considers depression-relevant, offering a bridge between statistical learning and clinical interpretability.

\subsubsection{Enhanced context with bigrams}
Individual words in social media posts may not always sufficiently reflect the author’s mental health state. However, when combined with other words, they can reveal depression-related signals. For example, the words \textit{pretending} and \textit{happy} are neutral or positive individually, but together as \textit{pretending happy}, they convey a negative tone. Similarly, \textit{feel} and \textit{nothing} are vague individually, but their combination suggests loss of pleasure, a common symptom of depression \cite{beck_ii}. To capture such patterns, we define bigram embeddings using a one-dimensional convolution operation:  
\begin{equation}
    \mathbf{b}_{i,j} \leftarrow \text{GELU} \left( \mathbf{W} \cdot 
    \begin{bmatrix}
        \mathbf{h}'_{i,j} \\
        \mathbf{h}'_{i,j+1}
    \end{bmatrix}
    + \mathbf{b} \right), \quad j = 1, \ldots, L-1
\end{equation}
where $\mathbf{h}'_{i,j} \in  \mathbf{H}'_i $ is the contextual unigram embedding at position $j$ of post $p_i$, and $\mathbf{W}$ and $\mathbf{b}$ are trainable convolution parameters. The overall bigram representation of post $p_i$ is then:
\begin{equation}
    \mathbf{B}_i \leftarrow \{\mathbf{b}_{i,j}\}_{j=1}^{L-1}.
\end{equation}


\subsubsection{Cross-attention for clinical relevance}
While unigram embeddings capture individual token information, certain bigrams often convey richer depression-related meaning. To model this, we apply a cross-attention mechanism over bigram representations $\mathbf{B}_i$, guided by domain knowledge from a depression-specific knowledge graph $\mathbf{G}_i$ (detailed later in Section \ref{sec:domainknowledgerepresentation}). Unlike standard self-attention, cross-attention allows the model to modulate bigram representations using external domain knowledge. Specifically, queries ($\mathbf{Q}$) are the bigram embeddings $\mathbf{B}_i$, representing the content of the social media post; keys ($\mathbf{K}$) are the knowledge graph embeddings $\mathbf{G}_i$, representing clinically relevant concepts; and values ($\mathbf{V}$) are the bigram embeddings $\mathbf{B}_i$, which are weighted according to their relevance to the knowledge graph. The cross-attention with multiple heads is computed as:  
\begin{align}
    \mathbf{\Omega}_i^k &\leftarrow \sigma \left( \frac{(\mathbf{B}_{i}\mathbf{\Theta}^k_{\mathbf{Q}}) (\mathbf{G}_{i}\mathbf{\Theta}^k_{\mathbf{K}})^T}{\sqrt{d_k}} \right)  (\mathbf{B}_{i}\mathbf{\Theta}^k_{\mathbf{V}}) \label{eqn:crossattention} \\
    \mathbf{B}'_i &\leftarrow \bigoplus_{k=1}^{\kappa_b} \left[ \mathbf{\Omega}_i^k \right] \mathbf{\Theta}_O \label{eqn:crossattention2}
\end{align} 
where $\mathbf{\Theta}^k_{\mathbf{Q}}$, $\mathbf{\Theta}^k_{\mathbf{K}}$, $\mathbf{\Theta}^k_{\mathbf{V}}$, and $\mathbf{\Theta}_O$ are trainable parameters, and $\kappa_b$ is the number of attention heads.

By attending to domain knowledge, this mechanism produces bigram representations $\mathbf{B}'_i$ that integrate both data-driven cues and clinically validated depression knowledge, allowing the model to focus on word pairs most indicative of depression severity. The resulting representations are both predictive and interpretable, highlighting clinically relevant phrases in social media posts.


\subsection{Domain knowledge representation} \label{sec:domainknowledgerepresentation}
\setcounter{section}{4} 
\subsubsection{Constructing domain knowledge graph}

\begin{figure*}[!ht]
\centering
\includegraphics[width=0.8\textwidth]{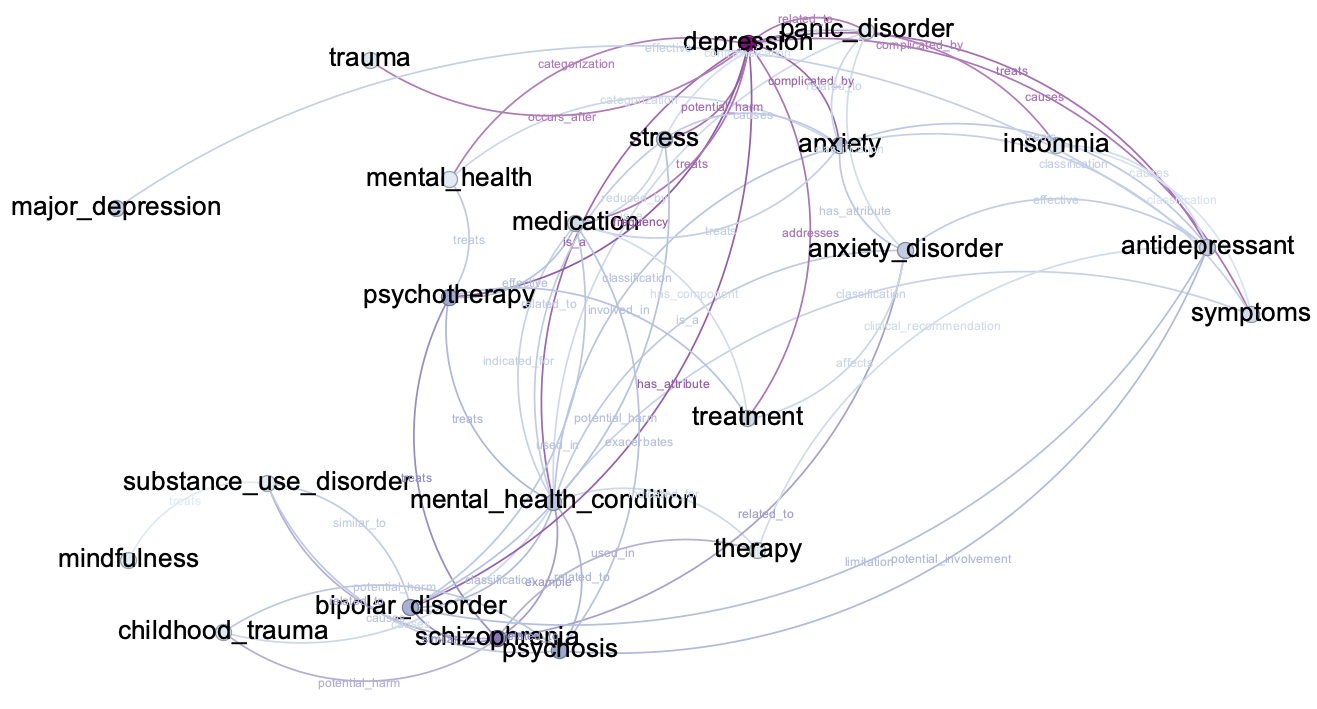}
\caption{Sample snapshot of the Mental Health Knowledge Graph (MHKG)}
\label{fig:kg_dat}
\end{figure*}

We construct a mental health knowledge graph (MHKG), denoted as $\mathcal{G}(\mathcal{V},\mathcal{E}, \mathcal{C}, \mathcal{F}, \mathcal{A})$, to capture clinically relevant mental health concepts and their relationships. Here, $\mathcal{V}$ represents nodes corresponding to mental health–related entities, $\mathcal{E}$ represents directed edges (relations), and $\mathcal{C}$ contains mental health context sentences. Each edge \( e_{s,t} \in \mathcal{E} \) is represented as a triplet \( \langle v_s, r_{s,t}, v_t \rangle \), where \( v_s \) and \( v_t \) are the source and target nodes, and \( r_{s,t} \) denotes the relation type. $\mathcal{F} \in \mathbb{R}^{|\mathcal{V}| \times 300}$ denotes the node embeddings of MHKG. Each node embedding $\mathbf{f}_k \in \mathcal{F}$ is computed by applying a pre-trained $\mathtt{FastText}$ model to the textual attribute of node \( v_k \), i.e., $\mathbf{f}_k = \mathtt{FastText}(v_k)$. Similarly, $\mathcal{A} \in \mathbb{R}^{|\mathcal{E}| \times 300}$ represents edge embeddings, where each edge embedding $\mathbf{a}_{s,t} \in \mathcal{A}$ is obtained by applying $\mathtt{FastText}$ to the textual attribute of the relation \( r_{s,t} \), i.e., $\mathbf{a}_{s,t} = \mathtt{FastText}(r_{s,t})$. These embeddings allow the model to encode both nodes and relations in a continuous vector space, preserving semantic and clinical relevance for downstream attention and representation learning. Figure~\ref{fig:kg_dat} shows a snapshot of our MHKG, illustrating mental health–related entities and their relationships. Its construction begins with generating a relevant knowledge base, which is followed by its validation, normalisation, and specific MHKG retrieval.

\noindent \textit{\textbf{Generating a knowledge base.}} In our study, the knowledge base is represented as a set of mental health–related contexts and their associated triplets. We first examine the REBEL dataset\footnote{https://huggingface.co/datasets/Babelscape/rebel-dataset} \cite{huguet-cabot-navigli-2021-rebel-relation}, a benchmark Wikipedia-sourced knowledge base containing over 3M contexts and triplets. However, as a general-purpose resource, REBEL lacks sufficient coverage of mental health–related contexts. To address this gap, we augment the dataset by extracting potential mental health–specific contexts and triplets leveraging Wikipedia and large language models (LLMs). For context collection, we employ the Wikipedia API\footnote{https://www.mediawiki.org/wiki/API:Main\_page} to scrape mental health–related content. In total, we collect information from 183 Wikipedia pages, divided into two categories: (i) 30 \textit{core topics}, consisting of names of mental disorders and related clinical protocols, and (ii) 153 \textit{secondary topics}, including subtypes, causes, and symptoms of these disorders. This ensures the contexts remain focused and clinically relevant while reducing sparsity. The full text of the core topic pages and summaries of the secondary topic pages are concatenated to form the final context corpus, consisting of 10,559 sentences. Tables~\ref{tab:core} and~\ref{tab:sec} list the core and secondary topics, respectively. Next, we apply a few-shot prompting strategy with $\mathtt{GPT-4o-mini}$\footnote{https://platform.openai.com/docs/models/gpt-4o-mini}, inspired by~\cite{zhang2024extract}, to extract triplets from the collected contexts. The exact prompt used is shown in Table~\ref{tab:prompts}. This process yields 10,559 samples, producing a total of 41,676 triplets.

\noindent \textit{\textbf{Validating clinical relevance.}}  
A triplet $\tau = \langle s, r, t\rangle \in \mathcal{E}$ is considered clinically relevant if it contains entities that are specific, meaningful, and clearly related to mental health (e.g., symptoms, disorders, emotions, behaviours, treatments, or psychological mechanisms). Furthermore, the relation $r$ between entities must represent a causal, plausible, and interpretable connection within the mental health domain. To operationalise this, we curate $\mathcal{M}$ as a predefined set of eligible entities derived from Tables~\ref{tab:core} and \ref{tab:sec}. A triplet $\tau$ is labelled clinically relevant if the following condition holds:  
\begin{equation}
\max\left(\max_{m \in \mathcal{M}} \varsigma(s, m), \; \max_{m \in \mathcal{M}} \varsigma(t, m) \right) > 0.9 \;\land\; \phi(\tau)
\label{eq:triplet_filter}
\end{equation}
where $\varsigma(\cdot,\cdot) \in [0,1]$ denotes a semantic similarity function, and $\phi(\tau) \in \{\text{True}, \text{False}\}$ is a triplet validity function implemented using an LLM (see Table~\ref{tab:prompts}). This condition ensures that each triplet is both semantically aligned with the mental health domain and validated as clinically meaningful. Applying this validation step yields 2,481 contexts and 6,433 validated triplets in total.

\noindent\textit{\textbf{Human-assisted normalisation.}} As the validated knowledge base contents are sourced from two different origins (REBEL and Wikipedia), this introduces heterogeneity. In addition, the newly generated triplets using LLMs may deviate in syntax, style, and phrasing, thereby increasing sparsity and inconsistency. For example, relation phrases such as $\mathtt{causes}$, $\mathtt{causeTo}$, $\mathtt{cause\_to}$, and $\mathtt{cause to}$ all share the same meaning but differ syntactically. They can therefore be interpreted as distinct relations. To address this, we perform triplet normalisation with the assistance of human annotators. Entities show similar inconsistencies as well. Therefore, we first normalise entities by identifying those with similar meanings using the semantic similarity of their embedding vectors. If entities with high similarity have the same or close meanings, a unique representative entity is selected. The same logic is applied to relation phrases. However, an additional sanity check is required, as lower cardinality of relation sets is generally maintained in benchmark knowledge graphs due to their higher predictability. For this purpose, we apply agglomerative clustering to the principal component vectors of their FastText embedding vectors, obtained via principal component analysis (PCA). The resulting clusters are then reviewed by human annotators, and 177 unique normalised relation types are finalised, as reported in Table~\ref{tab:normrels}. The final MHKG generated after normalisation contains 2,461 contexts and 6,371 triplets comprising 177 unique relations and 4,098 unique entities.

\noindent \textit{\textbf{Retrieving post-specific MHKG.}}  
As the overall MHKG contains a large amount of information, using it in its entirety for machine learning is both computationally expensive and unnecessary. Many triplets may also be irrelevant for a given post. To address this, we retrieve only a post-specific subgraph. For each post $p_i \in \mathcal{P}$, the retrieved list of triplets is defined as:
\begin{align}
    \mathcal{T}_i \; \leftarrow \bigcup_{c \in \text{TopK}_\mathcal{C}[\varsigma(p_i,\mathcal{C})]} \mathcal{T}(c)
\end{align}
where $\mathcal{C}$ denotes the set of all available context phrases, $\varsigma(\cdot,\cdot)$ is a semantic similarity function, and $\mathcal{T}(c)$ returns the list of triplets associated with context $c$. For computing semantic similarity, we adopt a pre-trained cross-encoder model from the \texttt{sentence\_transformers} library. The retrieval process then selects the top $K$ most similar contexts to the input post $p_i$ and collects all associated triplets. As a result, each post is paired with its own specific set of knowledge triplets, which are subsequently used to construct the corresponding MHKG embeddings.


\subsubsection{Knowledge graph representation learning}
To capture structural properties in our MHKG and extract clinically relevant signals for depression analysis, we employ residual sequential layers of the graph isomorphism network with edge features (GINE) \cite{xu2018powerful}. GINE is a graph neural network designed to capture complex structural patterns via edge features alongside node features, allowing it to distinguish graphs even with similar node attributes. \textsc{AttentionDep} aggregates information up to $n$-hop neighbours by stacking $n$ GINE layers, leveraging GINE's strong representational capacity. We process the retrieved MHKG for post $p_i$, denoted as $\mathcal{G}_i(\mathcal{V}_i,\mathcal{E}_i, \mathcal{C}_i, \mathcal{F}_i, \mathcal{A}_i)$, using $n$ consecutive GINE layers as defined in Equation~\ref{eqn:gin}. The embedding $\mathbf{\theta}_k^{(l)}$ is learned for each node $v_k \in \mathcal{V}$ at layer $l$, where $\mathbf{\theta}_k^{(0)} = \mathbf{f}_k$ is the initial node feature vector, $\epsilon$ denotes the relative importance of the target node compared to its neighbours, $\mathcal{N}(v_k)$ represents the neighbourhood of node $v_k$, and $\mathbf{a}_{k,u}$ represents the edge attributes between nodes $v_k$ and $v_u$.
\begin{align}
\mathbf{\theta}_k^{(l+1)} \leftarrow \mathbf{\theta}_k^{(l)} + \text{MLP}\Bigg(& (1 + \epsilon^{(l+1)}) \cdot \mathbf{\theta}_k^{(l)} \nonumber \\
&+ \sum_{v_u \in \mathcal{N}(v_k)} \left( \mathbf{\theta}_u^{(l)} + \mathbf{a}_{k,u} \right) \Bigg)
\label{eqn:gin}
\end{align}
The final MHKG representation for post $p_i$, which serves as the domain knowledge in Equation~\ref{eqn:crossattention}, is computed as:
\begin{equation} \label{eqn:maxpool}
    \mathbf{G}_i \leftarrow \{\mathbf{\theta}_v^{(n)}\}_{v = 1}^{|\mathcal{V}|}.
\end{equation}

\subsection{Depression severity with ordinal representation}
\subsubsection{Depression severity classification}
After obtaining the domain knowledge–fused representation from Equations~\ref{eqn:crossattention} and~\ref{eqn:crossattention2}, we proceed to the final step of depression severity estimation. The prediction for post $p_i$ is generated as:
\begin{align}
\hat{y}_i &\leftarrow \sigma\left( \text{MLP}\left( \frac{1}{\|\mathbf{M}_i\|_1} \mathbf{1}^T (\mathbf{M}_i \odot \mathbf{B}'_{i})\right)\right) \\
f(p_i) &\leftarrow \text{argmax} \; \hat{y}_i  
\label{eqn:logits}
\end{align}
where, $\mathbf{M}_i \in \{0,1\}^{L \times 1}$ is a binary masking matrix indicating whether a representation corresponds to a valid token ($1$) or padding ($0$). This ensures that padding tokens do not influence the decision-making process. The element-wise multiplication $\mathbf{M}_i \odot \mathbf{B}'_{i}$ followed by averaging is called masked mean pooling, which aggregates information from only valid tokens, and $\text{MLP}(\cdot)$ denotes a feed-forward multilayer perceptron network. Finally, $f(p_i)$ represents the predicted depression severity for post $p_i$.



\subsubsection{Ordinal encoding and backpropagation}
Given the ordinal nature of depression severity levels, we adopt ordinal regression inspired by Sawhney et al. \cite{sawhney2021towards}. Let $\mathcal{Y} = \{\text{minimum}=0, \text{mild}=1, \text{moderate}=2, \text{severe}=3\}$ denote the label space. For a post $p_i$ with true severity $y_i \in \mathcal{Y}$, we generate a soft label distribution $\mathcal{P}^i = [\rho_0^i, \rho_1^i, \rho_2^i, \rho_3^i]$ as follows:
\begin{align}
\rho_j^i &\leftarrow \frac{\exp(-\phi(y_i,y_j))}{\sum_{y_k \in \mathcal{Y}} \exp(-\phi(y_i,y_k))} \label{eqn:softlabel} \\
\phi(y_i,y_j) &\leftarrow \beta |y_i - y_j| \label{eqn:cost}
\end{align}
where, $\phi(y_i,y_j)$ is a cost function capturing the distance between the true severity $y_i$ and each severity level $y_j \in \mathcal{Y}$, and $\beta$ is a hyperparameter controlling the penalty magnitude for mispredictions. Larger differences between $y_i$ and $y_j$ result in lower probabilities $\rho_j^i$, reflecting the ordinal structure of the labels. The prediction error for post $p_i$ is measured using cross-entropy loss:
\begin{align}
\mathcal{L}_i \leftarrow \sum_{j \in \mathcal{Y}} y_j^i \log(\hat{y}_j^i)
\end{align}
This loss is minimised during training to update the model parameters and learn an ordinal-aware mapping from the input representation to depression severity levels.

\section{Experiments and Results} \label{sec:experimentalresults}

\subsection{Datasets}

We construct three experimental datasets from Reddit. Curated in collaboration with domain experts and aligned with DSM criteria \cite{apa2013diagnostic}, two datasets support multi-class severity-level analysis, while the third enables binary depression detection. Although our analysis focuses on Reddit, the embedding and modelling framework is adaptable to other platforms and languages.

To build these datasets, we start from two publicly available Reddit datasets. The first \cite{naseem2022early} is an augmented version of the Dreaddit dataset \cite{turcan-mckeown-2019-dreaddit} refined according to DSM-V standards. The second \cite{sampath2022data}, is collected via the Reddit API from mental health–related subreddits. While both provide valuable content and labels, they are not directly used in experiments due to limitations in class coverage and balance. We combine and refine posts from these sources to create our primary experimental dataset, $\mathbf{D_4}$, supporting four-class severity classification. As the \textit{‘mild’} class is underrepresented (fewer than 300 posts), we apply augmentation techniques to increase its size. For experiments with a reduced-class setting, we derive a three-class subset, $\mathbf{D_3}$, by excluding mild posts. Binary depression detection is performed using $\mathbf{D_2}$ \cite{pirina-coltekin-2018-identifying}.

\begin{table}[h!]
\centering
\caption{Summary of datasets used for evaluation.}
\begin{tabular}{lcp{4cm}c}
\toprule
Dataset & Classes & Class distribution & Source\\
\midrule
$\mathbf{D_4}$ & 4 & Minimum: 1500, Mild: 580, Moderate: 2000, Severe: 1000 & Reddit \\
$\mathbf{D_3}$ & 3 & Minimum: 1500, Moderate: 2000, Severe: 1000 & Reddit \\
$\mathbf{D_2}$ & 2 & 1293 depressed, 548 control & Reddit \\
\bottomrule
\end{tabular}
\label{tab:datasets}
\end{table}

\subsection{Experimental Setup}

We select optimal hyperparameters based on the highest graded $\text{F}_1$ score using the Optuna framework\footnote{\url{https://optuna.org}} with the Tree-structured Parzen Estimator (TPE) sampler. The chosen hyperparameters are: learning rates of $9.8\times10^{-5}$ and $5.4\times10^{-5}$ for \textbf{D3} and \textbf{D4}, respectively; 200 training epochs; 4 and 2 unigram attention heads for \textbf{D3} and \textbf{D4}; 4 bigram attention heads for both datasets; a dropout rate of 0.3; batch size of 128; severity scale of 4.5; hidden size of 128; 2 LSTM layers; 3 and 1 GNN layers for \textbf{D3} and \textbf{D4}; and a maximum input length of 256 tokens. We fine-tune language models using the Huggingface Transformers library and implement\footnote{The code and data will be publicly available upon acceptance.} all experiments in PyTorch 2.1 with the Adam optimizer. Experiments run on the Viking HPC Cluster at the University of York, equipped with 12,864 CPU cores, 512 GB RAM, and NVIDIA H100 GPUs.

\subsection{Evaluation Metrics}
Standard classification metrics such as Precision, Recall, and $F_1$ score treat classes as independent and ignore the ordinal nature of labels, making them less informative for severity-level prediction. To address this, we employ Graded Precision (GP), Graded Recall (GR), and Graded $F_1$ (GF) scores \cite{10.1145/3308558.3313698,sawhney2021towards}, which account for the ordinal structure of depression severity levels. These metrics redefine False Negatives (FN) and False Positives (FP) to reflect whether a prediction underestimates or overestimates the true severity:
\begin{align}
    \text{FN} = \frac{1}{N_T} \sum_{i=1}^{N_T} \mathbb{I}(k_a^i > k_p^i), \;\;\;
    \text{FP} = \frac{1}{N_T} \sum_{i=1}^{N_T} \mathbb{I}(k_p^i > k_a^i)
\end{align}
where, $k_p^i$ and $k_a^i$ denote the predicted and actual severity levels, respectively. A prediction is considered a false negative if it underestimates the true severity ($k_p < k_a$) and a false positive if it overestimates it ($k_p > k_a$). True positives occur when the predicted and actual severity levels match exactly.




\subsection{Baseline Models}

We compare \textsc{AttentionDep} against twelve baselines, including six generic neural architectures and six state-of-the-art (SOTA) depression detection models.  

\noindent \textit{Generic neural baselines.}  
\begin{itemize}
    \item \textbf{LSTM, GRU, BiLSTM, BiGRU}: Recurrent networks for sequential data modelling.
    \item \textbf{CNN}: Extracts salient local features through convolution, widely used in NLP tasks.
    \item \textbf{BERT}: Pre-trained transformer model for text representation.
\end{itemize}

\noindent \textit{SOTA models.}  
\begin{itemize}
    \item \textbf{DepressionNet} \cite{zogan2021depressionnet}: Performs binary depression classification using text summarisation, BERT, BiGRU, attention, and social media features.
    \item \textbf{Naseem et al.} \cite{naseem2022early}: Utilises TextGCN, BiLSTM, and attention for ordinal regression of depression severity.
    \item \textbf{HAN} \cite{yang-etal-2016-hierarchical}: Hierarchical attention network that models word- and sentence-level structures.
    \item \textbf{HCN} and \textbf{HCN+} \cite{10041797}: Hierarchical convolutional attention networks for word- and tweet-level feature extraction; \textbf{HCN+} includes an additional MLP layer.
    \item \textbf{\textsc{DepressionX}} \cite{yusif2025depressionx}: Our previous model with a fixed knowledge graph and concatenation-based for knowledge infusion.
\end{itemize}

\subsection{Performance Comparison}

Table~\ref{tab:baseline_res} shows results across all datasets. All experiments are run five times and the average evaluation measures and their standard deviations are reported. Our model, \textsc{AttentionDep}, consistently outperforms all compared models, achieving substantial improvements in both severity-level and binary depression detection. 

\begin{table}[ht]
\centering
\caption{Performance comparison of the proposed AttentionDep across datasets D4, D3, and D2. Results are reported as mean ± standard deviation for graded F1 (GF), graded precision (GP), and graded recall (GR).}
\scalebox{0.82}{
\begin{tabular}{lccc}
\toprule
\textbf{Model} & \textbf{GF} & \textbf{GP} & \textbf{GR} \\
\midrule
\multicolumn{4}{c}{\textbf{D4}} \\
\midrule
CNN & 0.7419 ± 0.0104 & 0.7826 ± 0.0126 & 0.7052 ± 0.0114\\
LSTM & 0.7225 ± 0.0140  & 0.7477 ± 0.0193 & 0.6992 ± 0.0149\\
GRU & 0.7346 ± 0.0150 & 0.7654 ±0.0172 & 0.7064 ± 0.0185\\
BiLSTM & 0.7287 ± 0.0161 &  0.7483 ± 0.0229 &  0.7113 ± 0.0304 \\
BiGRU & 0.7373 ± 0.0116 & 0.7686 ± 0.0129 & 0.7093 ± 0.0268\\
BERT & 0.7241 ± 0.0134 & 0.7521 ± 0.0236  & 0.6993 ± 0.0261 \\
DepressionNet \cite{zogan2021depressionnet} & 0.6994 ± 0.0149  & 0.7257 ± 0.0329 & 0.6760 ± 0.0137 \\
Naseem et al. \cite{naseem2022early} & 0.5663 ± 0.0323  & 0.4927 ± 0.0124 & 0.6657 ± 0.0345 \\
HAN \cite{yang-etal-2016-hierarchical} & 0.7250 ± 0.0118 & 0.7693 ± 0.0337 & 0.6892 ± 0.0443 \\
HCN \cite{10041797} & 0.6898 ± 0.0352 & 0.7223 ± 0.0351 & 0.6628 ± 0.0374 \\
HCN+ \cite{10041797} &  0.7032 ± 0.0097 & 0.7269 ± 0.0337 & 0.6825 ± 0.0174 \\
DepressionX \cite{yusif2025depressionx} & 0.4838 ± 0.1475 & 0.4332 ± 0.0862 & 0.5677 ± 0.2101\\
\textbf{AttentionDep} & \textbf{0.7952 ± 0.0133} & \textbf{0.7805 ± 0.0185} & \textbf{0.8107 ± 0.0120} \\
\midrule
\multicolumn{4}{c}{\textbf{D3}} \\
\midrule
CNN & 0.7525 ± 0.0126 & 0.7985 ± 0.0170 & 0.7116 ± 0.0129\\
LSTM & 0.7342 ± 0.0203 & 0.7555 ± 0.0380 & 0.7179 ± 0.0479 \\
GRU & 0.7400 ± 0.0218 & 0.7501 ± 0.0612 & 0.7404 ± 0.0676\\
BiGRU & 0.7369  ± 0.0185 & 0.7693 ± 0.0468 &0.7141 ± 0.0602\\
BiLSTM & 0.7369  ± 0.0145 &  0.7571 ± 0.0385 & 0.7205  ± 0.0348\\
BERT & 0.7461 ± 0.0117 & 0.7846 ± 0.0172 & 0.7125 ± 0.0311 \\
DepressionNet \cite{zogan2021depressionnet} & 0.7223 ± 0.0151  & 0.7551 ± 0.0392 & 0.6940 ± 0.0211 \\
Naseem et al. \cite{naseem2022early} & 0.6198 ± 0.0081  & 0.5838 ± 0.0220 & 0.6614 ± 0.0086 \\
HAN \cite{yang-etal-2016-hierarchical} & 0.7040 ± 0.0303 & 0.7682 ± 0.0636 & 0.6544 ± 0.0485 \\
HCN \cite{10041797} & 0.6597 ± 0.0040 &  0.7589 ± 0.0325 & 0.5843 ± 0.0162 \\
HCN+ \cite{10041797} & 0.6817 ± 0.0273 & 0.7925 ± 0.0313 & 0.5986 ± 0.0295 \\
DepressionX \cite{yusif2025depressionx} & 0.5301 ± 0.0934 & 0.5808 ± 0.2444 &  0.6552 ± 0.2214\\
\textbf{AttentionDep} & \textbf{0.8052 ± 0.0095} & \textbf{0.7732 ± 0.0195} & \textbf{0.8404 ± 0.0239} \\
\midrule
\multicolumn{4}{c}{\textbf{D2}} \\
\midrule
CNN & 0.8745 ± 0.0185 & 0.8755 ± 0.0178 & 0.8772 ± 0.0171 \\
LSTM &  0.8830 ± 0.0132 & 0.8856 ± 0.0132 & 0.8821 ± 0.0129 \\
GRU & 0.8906 ± 0.0117 & 0.8939 ± 0.0116 & 0.8897 ± 0.0124 \\
Bi-GRU & 0.8825 ± 0.0120 & 0.8853 ± 0.0100 & 0.8821 ± 0.0135\\
Bi-LSTM & 0.8751 ± 0.0245  & 0.8787 ± 0.0229 & 0.8740 ±  0.0255 \\
BERT & 0.8508 ± 0.0222 & 0.8533 ± 0.0209 & 0.8561 ± 0.0201 \\
DepressionNet \cite{zogan2021depressionnet} & 0.8183 ± 0.0193  & 0.8190 ± 0.0196 & 0.8191 ± 0.0196 \\
Naseem et al. \cite{naseem2022early} & 0.8251 ± 0.0009  & 0.7823 ± 0.0012 & 0.8601 ± 0.0032 \\ 
HAN \cite{yang-etal-2016-hierarchical} & 0.8620 ± 0.0122 & 0.8633 ± 0.0135 & 0.8613 ± 0.0123 \\
HCN \cite{10041797} & 0.8253 ± 0.0331 & 0.8311 ± 0.0300 & 0.8208 ± 0.0374 \\
HCN+ \cite{10041797} & 0.8265 ± 0.0275 & 0.8258 ± 0.0277 & 0.8274 ± 0.0280\\
DepressionX \cite{yusif2025depressionx} & 0.8015 ± 0.0407 & 0.7505 ± 0.1094 & 0.9117 ± 0.1599 \\
\textbf{AttentionDep} & \textbf{0.9187 ± 0.0140} & \textbf{0.9195 ± 0.0140} & \textbf{0.9185 ± 0.0140} \\
\bottomrule
\end{tabular}
}
\label{tab:baseline_res}
\end{table}

Among the generic neural baselines, RNN-based models (LSTM, BiLSTM, GRU, BiGRU) perform reasonably well due to their ability to capture sequential dependencies, but are limited by vanishing gradients. CNNs perform slightly better by detecting salient local patterns, while BERT achieves stronger results by leveraging transformer-based contextual representations, particularly on \textbf{D3}, where ordinal distinctions are more prominent. 

The specialised models - DepressionNet \cite{zogan2021depressionnet}, HAN \cite{yang-etal-2016-hierarchical}, HCN \cite{10041797}, HCN+ \cite{10041797}, Naseem et al.~\cite{naseem2022early} and DepressionX \cite{yusif2025depressionx} - fall short of expectations despite their architectural complexity. Their weaker performance is likely due to insufficient integration of mental-health-specific linguistic cues, which limits their capacity to model fine-grained severity distinctions. Our previous model, \textsc{DepressionX}, also underperforms because of its simplistic knowledge integration strategy and reliance on evaluation settings that are not well aligned with severity classification.

In contrast, \textsc{AttentionDep} achieves over 5\% higher graded $\text{F}_1$ than the best baseline on both severity datasets, reaching 80.5\% on \textbf{D3} (with 77.3\% precision and 84.0\% recall) and 78.5\% on \textbf{D4} (78.0\% precision, 81.1\% recall). Performance is slightly stronger on \textbf{D3}, likely because the \textit{mild} class in \textbf{D4} overlaps semantically with neighbouring categories, making classification harder. Finally, evaluation on the binary dataset \textbf{D2} further confirms the robustness of our approach, where \textsc{AttentionDep} outperforms all baselines by a clear margin.

\subsection{Ablation Study}

We conduct an ablation study to evaluate the contribution of each component in \textsc{AttentionDep}, as shown in Table~\ref{tab:ablation_res}. All experiments are run five times and the average evaluation measures and their standard deviations are reported. The six configurations are: 
\begin{itemize}
\item[-] \textbf{A0}: Unigram-only features; 
\item[-]\textbf{A1}: A0 with concatenated KG representation; 
\item[-]\textbf{A2}: A0 with cross-attention-based KG integration; 
\item[-]\textbf{A3}: Unigram + bigram hierarchical features; 
\item[-]\textbf{A4}: A3 with concatenated KG; 
\item[-]\textbf{A5}: Full model (unigram + bigram + cross-attention-based KG). 
\end{itemize}

\begin{table*}[ht]
\centering
\caption{Ablation study results showing graded evaluation measures (GF, GP, and GR; Mean ± Std) across D4, D3, and D2 for different model configurations.}
\begin{tabular}{llccc}
\toprule
Dataset & Configuration & GF & GP & GR \\
\midrule
\multirow{6}{*}{\textbf{D4}} 
& \textbf{A0}: Unigrams only                 & 0.7408 ± 0.0164 & 0.7493 ± 0.0081 & 0.7334 ± 0.0329 \\
& \textbf{A1}: Unigarms +  KG (concat) & 0.7637 ± 0.0143 &  0.7710 ± 0.0196 & 0.7566 ± 0.0117\\
& \textbf{A2}: Unigrams + KG (cross att.) &  0.7679 ± 0.0056 & 0.7801 ± 0.0196 & 0.7566 ± 0.0134 \\
& \textbf{A3}: Unigrams + Bigrams           & 0.7595 ± 0.0137 & 0.7548 ± 0.0226 & 0.7646 ± 0.0129 \\
& \textbf{A4}: Unigrams + Bigrams +  KG (concat)   &  0.7642 ± 0.0085 & 0.7500 ± 0.0074 & 0.7794 ± 0.0200 \\
& \textbf{A5}:\textbf{ Full:} Unigrams + Bigrams +  KG (cross att.)  & \textbf{0.7952 ± 0.0133} & \textbf{0.7805 ± 0.0185} & \textbf{0.8107 ± 0.0120} \\
\midrule
\multirow{6}{*}{\textbf{D3}} 
& \textbf{A0}: Unigrams only                 & 0.7730 ± 0.0131 & 0.7622 ± 0.0083 & 0.7927 ± 0.0226 \\
& \textbf{A1}: Unigarms +  KG (concat) & 0.7810 ± 0.0138 & 0.7698 ± 0.0226 & 0.7930 ± 0.0138 \\
& \textbf{A2}: Unigrams + KG (cross att.) & 0.7901 ± 0.0219  & 0.7783 ± 0.0116 & 0.8025 ± 0.0332 \\
& \textbf{A3}: Unigrams + Bigrams           & 0.7878 ± 0.0115 & 0.7559 ± 0.0197 & 0.8229 ± 0.0142 \\
& \textbf{A4}: Unigrams + Bigrams +  KG (concat)   & 0.7934 ± 0.0122 & 0.7623 ± 0.0195 & 0.8277 ± 0.0189 \\
& \textbf{A5}: \textbf{Full:} Unigrams + Bigrams +  KG (cross att.)  & \textbf{0.8052 ± 0.0095} & \textbf{0.7732 ± 0.0195} & \textbf{0.8404 ± 0.0239} \\
\midrule
\multirow{6}{*}{\textbf{D2}} 
& \textbf{A0}: Unigrams only                & 0.8800 ± 0.0096 & 0.8719 ± 0.0090 & 0.8900 ± 0.0125 \\
& \textbf{A1}: Unigarms +  KG (concat) & 0.8990 ± 0.0114 & 0.9013 ± 0.0121 & 0.8984 ± 0.0115 \\
& \textbf{A2}: Unigrams + KG (cross att.)               & 0.9009 ± 0.0236 &  0.9039 ± 0.0226 & 0.9001 ± 0.0239 \\
& \textbf{A3}: Unigrams + Bigrams           & 0.8866 ± 0.0089 & 0.8819 ± 0.0153 & 0.8938 ± 0.0038 \\
& \textbf{A4}: Unigrams + Bigrams +  KG (concat)  & 0.8946 ± 0.0223 & 0.8865 ± 0.0242 & 0.9054 ± 0.0179 \\
& \textbf{A5}: \textbf{Full:} Unigrams + Bigrams +  KG (cross att.)   & \textbf{0.9187 ± 0.0140} & \textbf{0.9195 ± 0.0140} & \textbf{0.9185 ± 0.0140} \\
\bottomrule
\end{tabular}
\label{tab:ablation_res}
\end{table*}

The objectives of this ablation study are mainly to assess the impact of cross-attention-based knowledge infusion and bigram-level hierarchical context modelling. Unigram features alone provide notable performance for depression severity classification. Adding bigram features slightly enhances performance on severity datasets (\textbf{D4} and \textbf{D3}), consistent with our goal of improving interpretability rather than raw performance through bigram inclusion. Incorporating domain knowledge via MHKG consistently improves results. The cross-attention-based integration outperforms concatenation because it dynamically aligns textual and knowledge features. The full model achieves the best performance across both multi-class (\textbf{D4}, \textbf{D3}) and binary (\textbf{D2}) settings. The consistently low standard deviations demonstrate stability and robust convergence across configurations.

\subsection{Parametric Analysis}
We perform a parametric analysis using 100 hyperparameter tuning trials. To identify the most influential continuous parameters, we apply Spearman’s rank correlation coefficient ($\rho$). For categorical parameters, we use the Kruskal–Wallis (KW) test. The results are reported in Table~\ref{tab:hp_stats_d3_d4}. At the 0.05 significance level, the statistical tests indicate that the learning rate, number of LSTM layers, number of attention heads for both unigram and bigram representations, and the number of consecutive GNN layers are the most influential hyperparameters during tuning.

\begin{table}[!h]
\centering
\caption{Statistical tests for hyperparameter tuning. Significant $p$-values ($<$0.05) are in \textbf{bold}.}
\begin{tabular}{lccc}
\toprule
\textbf{Hyperparameter} & \textbf{Test} & \textbf{D3 (stat, $p$)} & \textbf{D4 (stat, $p$)}\\
\midrule
learning rate:\;$\eta$ & S      & 0.228, \textbf{0.0225}   & -0.485, \textbf{0.0000} \\
dropout rate:\;$d_p$        & S      & -0.062, 0.5410  & -0.020, 0.8458 \\
pen. scale:\;$\beta$          & S      & 0.025, 0.8063   & 0.188, 0.0613 \\
hidden size:\;$h$    & KW & 0.021, 0.8845  & 3.406, 0.0649 \\
$\#$ uni. att. heads:\;$\kappa_u$          & KW & 17.398, \textbf{0.0002} & 29.152, \textbf{0.0000} \\
$\#$ LSTM layers:\;$\delta$   & KW & 9.645, \textbf{0.0080}  & 15.504, \textbf{0.0004} \\
$\#$ bi. att. heads:\;$\kappa_b$      & KW & 6.848, \textbf{0.0326}  & 14.294, \textbf{0.0008} \\
$\#$ GNN layers:\; $n$    & KW & 9.877, \textbf{0.0072}  & 20.441, \textbf{0.0000}\\
\bottomrule
\end{tabular}
\label{tab:hp_stats_d3_d4}
\end{table}

\begin{figure*}[!h]
\begin{center}
 \includegraphics[width=\textwidth]{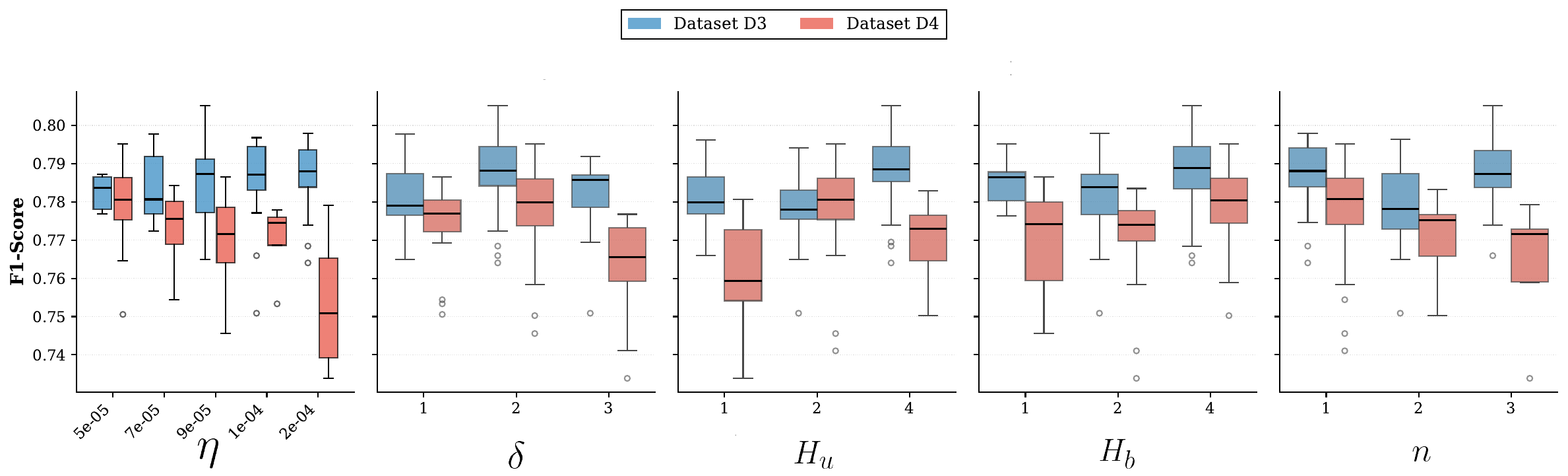}
 \caption{Effects of hyperparameters on graded F1 score for severity datasets D4 and D3.}
 \label{fig:param_anal}
\end{center}
\end{figure*}

Figure~\ref{fig:param_anal} shows the effects of varying hyperparameters on classification performance for the severity datasets.

\noindent \textit{\textbf{Learning rate.}} It regulates the step size of parameter updates. Low values lead to slow convergence or getting trapped in local minima, while excessively high values overshoot optimal minima, causing unstable learning. On \textbf{D4}, F1 decreases as the learning rate increases, indicating that high rates may skip information required to distinguish closely related categories (e.g., \textit{minimum}, \textit{mild}, \textit{moderate}). On \textbf{D3}, which excludes the mild class, a slight improvement occurs with higher rates, suggesting that reduced semantic overlap allows faster convergence.

\noindent\textit{\textbf{Number of LSTM layers.}} It controls the model’s ability to capture long-term dependencies. Two layers yield the best performance on both datasets. In contrast, the number of GNN layers, which governs how deeply the model integrates structural information from the knowledge graph, minimally impacts \textbf{D3} but strongly affects \textbf{D4}. Optimal settings are a single layer for \textbf{D4} and three layers for \textbf{D3}.

\noindent\textit{\textbf{Number of attention heads}}. For unigram and bigram layers, this number is highly influential, enhancing the model’s ability to capture diverse patterns such as emotional cues, symptoms, or potential causes simultaneously. For unigram representations, two heads are optimal for D4 and four for D3. For bigram representations, four heads are optimal for both datasets, highlighting the increased capacity of bigram encoding to capture depression-related patterns.

\subsection{Explainability Analysis}
Attention mechanisms can be associated with model explanation in generating responsible outputs, particularly in healthcare applications where the black-box nature of models raises ethical concerns. Feature importance is one of the most widely used model explanation techniques. Although the consensus is weak among scholars regarding whether attention weights represent feature importance, some studies have empirically validated this association~\cite{wiegreffe-pinter-2019-attention}. Mathematically, the partial derivative of the model output $y$ with respect to a feature $x_j$ is scaled by that feature’s attention weight $\alpha_j$, providing a basis to link attention to feature importance.

\begin{figure*}[!h]
\centering
\begin{subfigure}[b]{0.45\textwidth}
    \includegraphics[width=\textwidth]{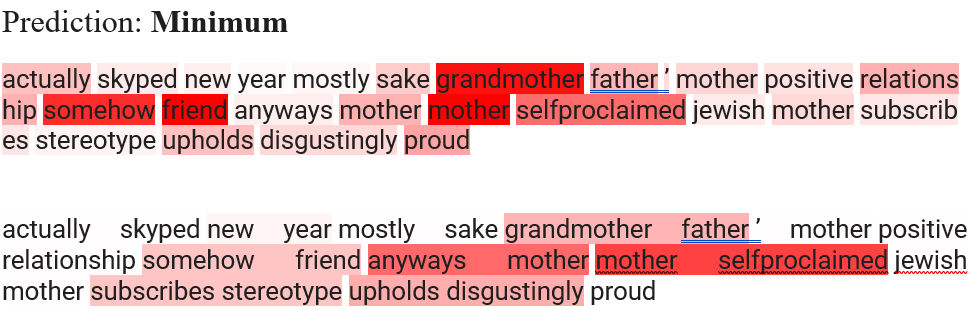}
    \caption{Minimum Depression Example}
    \label{fig:min}
\end{subfigure}
\hspace{0.02\textwidth}
\begin{subfigure}[b]{0.45\textwidth}
    \includegraphics[width=\textwidth]{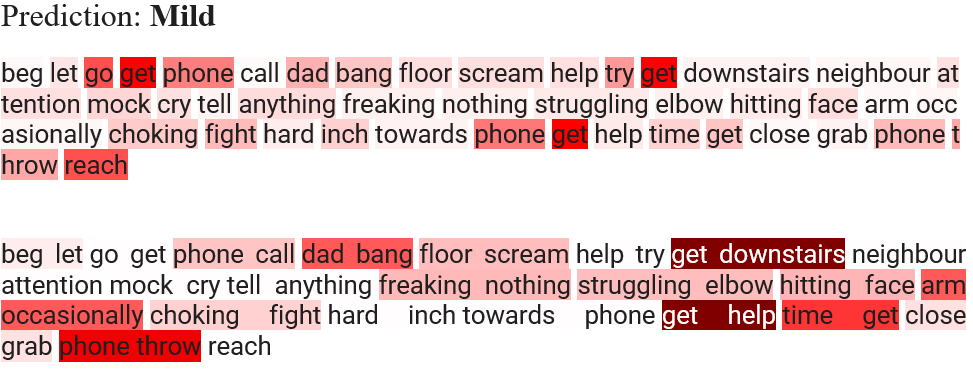}
    \caption{Mild Depression Example}
    \label{fig:mild}
\end{subfigure}

\vspace{0.15cm}

\begin{subfigure}[b]{0.45\textwidth}
    \includegraphics[width=\textwidth]{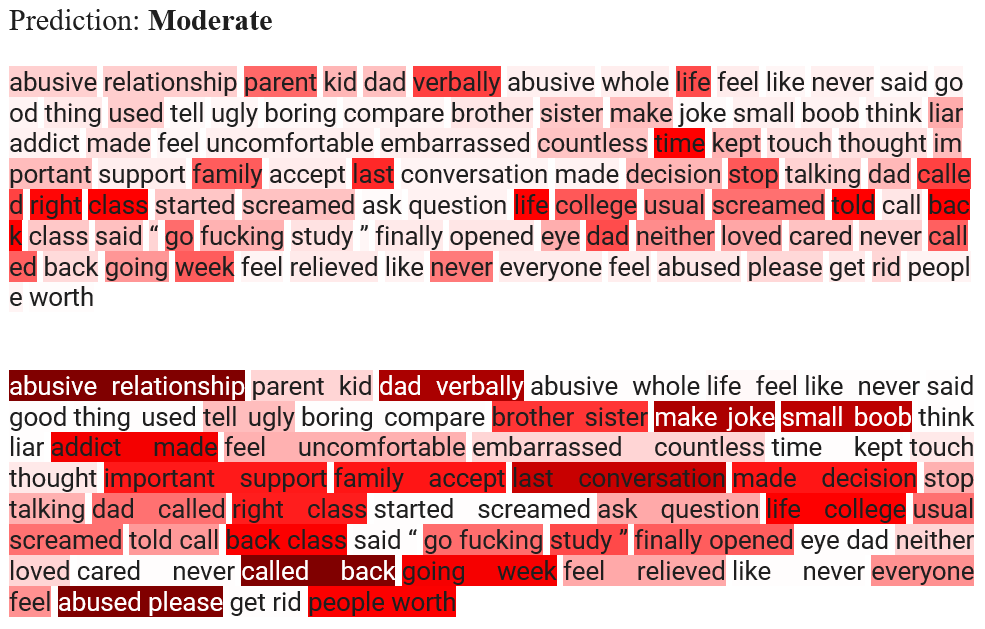}
    \caption{Moderate Depression Example}
    \label{fig:moderate}
\end{subfigure}
\hspace{0.02\textwidth}
\begin{subfigure}[b]{0.45\textwidth}
    \includegraphics[width=\textwidth]{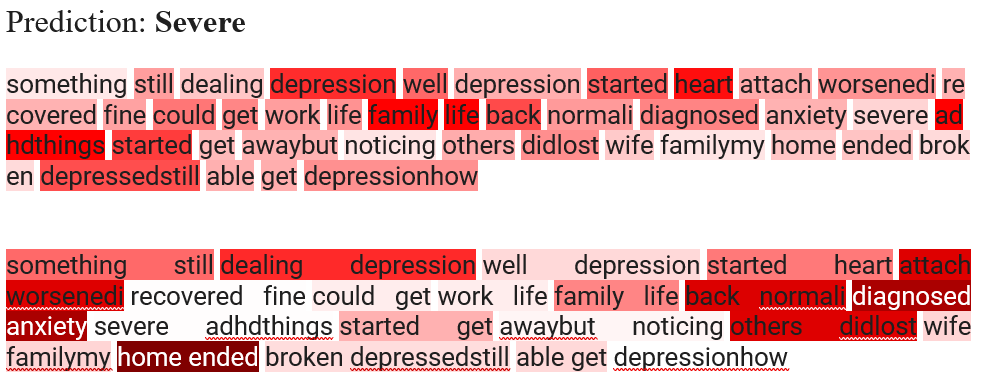}
    \caption{Severe Depression Example}
    \label{fig:severe}
\end{subfigure}

\caption{Token-level importance visualisations for representative posts across different depression severity levels. Hierarchical attention highlights unigrams and bigrams most influential in the model’s predictions, providing insight into the decision-making process.}
\label{fig:depression_examples}
\end{figure*}

We employ a hierarchical attention mechanism over unigram and bigram features to highlight salient terms, enabling us to analyse how the model makes decisions regarding depression severity. Figure~\ref{fig:depression_examples} presents representative input samples along with their most influential unigrams and bigrams. 

For the \textit{minimum} level of depression, the model predominantly focuses on non-depressive words; however, certain negative terms such as \textit{stereotype} or \textit{disgustingly} still receive attention. A similar pattern occurs in the \textit{mild} depression level, supporting the observation that the model struggles to distinguish mild depressive cues from non-depressive content. 

For \textit{moderate} and \textit{severe} depression, the utility of bigram-level attention becomes more pronounced. In the moderate case, individual words such as \textit{abusive} and \textit{relationship} hold moderate importance, but their combination as the phrase \textit{abusive relationship} significantly increases their saliency, exerting a stronger influence on the final decision. A similar phenomenon appears in the severe class, where phrases like \textit{diagnosed anxiety} receive greater attention than the individual words, highlighting critical patterns indicative of higher depression severity.

\section{Conclusion} \label{sec:conclusion}

In this paper, we present \textsc{AttentionDep}, a novel knowledge-infused and explainable model for predicting depression severity from social media text. \textsc{AttentionDep} integrates hierarchical attention mechanisms over unigram and bigram representations with external domain knowledge from a curated mental health knowledge graph (MHKG). This enables the identification of salient textual patterns relevant to different depression severity levels. Posts are encoded using FastText embeddings and a bidirectional LSTM to capture contextual dependencies, while convolutional operations extract bigram-level features. In parallel, knowledge from Wikipedia articles is used to construct an MHKG, which is processed via the GINE model to generate enriched knowledge representations. Cross-attention is then applied to fuse textual and MHKG-based features, producing comprehensive post representations for depression severity classification.

We evaluate \textsc{AttentionDep} on three datasets: \textbf{D4} and \textbf{D3} (multi-class severity) and \textbf{D2} (binary classification). Across all datasets, \textsc{AttentionDep} consistently outperforms baseline methods, achieving graded $\text{F}_1$ scores of 79.52\% on \textbf{D4} and 80.52\% on \textbf{D3}, exceeding all compared models by over 5\%. On the binary dataset \textbf{D2}, the model achieves a graded $\text{F}_1$ score of 91.87\%. These results highlight the effectiveness of combining domain knowledge with explainable deep learning for mental health prediction.

Despite these promising results, several limitations remain. While attention mechanisms provide interpretability and the model integrates clinically validated domain knowledge, future work could explore additional strategies to further quantify how attention weights correspond to specific depression-relevant features. Moreover, the current model focuses exclusively on textual data, omitting other potentially informative multimodal signals, such as behavioural features or social interactions. Future work could enhance interpretability and predictive performance by incorporating such signals and modelling user connections within social networks. Overall, \textsc{AttentionDep} demonstrates a robust framework for explainable, knowledge-driven depression detection from social media, offering valuable insights for AI-assisted mental health applications.

\textbf{Ethics Approval:} All analyses are conducted under strict ethical guidelines. This study is approved by the \textit{Physical Sciences Ethics Committee of the University of York} under application reference \textit{Ibrahimov20230330}. No personally identifiable information was accessed during this research.

\section*{References}

\bibliographystyle{unsrt}
\bibliography{sample-base}

\appendices
\section{Auxiliary Tables} \label{sec:app}

\begin{table*}[t]
\centering
\caption{LLM Prompt for Semantic Graph Construction} \label{tab:prompts}
\renewcommand{\arraystretch}{1.3}
\begin{tabular}{p{0.25\textwidth}|p{0.68\textwidth}}
\hline
\textbf{Purpose} & \textbf{Prompt} \\
\hline
Triplet Extraction & 
\ttfamily
Your task is to transform the given text into a mental health related semantic graph in the form of a list of triples. The triples must be in the form of [Entity1, Relationship, Entity2]. In your answer, please strictly only include the triples and do not include any explanation or apologies. Keep the entities and relations as simple and short as possible, and do not make them long, if it is not necessary.

Here are some examples:\newline
\{few\_shot\_examples\}

Now please extract triplets from the following text.\newline
Text: \{input\_text\} \\
\hline
Triplet Validation & \ttfamily
You are given a knowledge triplet in the form [subject, relation, object].
Determine if this triplet expresses a valid and meaningful relationship in the context of mental health.
A valid relation is one that expresses a clear, interpretable, and contextually appropriate connection—such as causal, descriptive, indicative, or therapeutic—between two mental health-relevant concepts.
Evaluate the triplet based on:

The subject and object must both be specific, meaningful, and clearly related to mental health (e.g., symptoms, disorders, emotions, behaviors, treatments, or psychological mechanisms). Avoid vague, overly broad, or overly technical terms unless they directly contribute to mental health understanding.

The relation must express a plausible and interpretable connection within the mental health domain.

Return only: yes or no.

Example (invalid due to vague object):
["models", "explain", "link between altered brain function and schizophrenia"] → no \\
\bottomrule
\end{tabular}

\end{table*}


\begin{table*}[!h]
\centering
\caption{Core mental health topics used in the knowledge graph}\label{tab:core}
\begin{tabularx}{\textwidth}{>{\raggedright\arraybackslash}X >{\raggedright\arraybackslash}X >{\raggedright\arraybackslash}X >{\raggedright\arraybackslash}X}
\toprule
Mental health & Mental disorder & Psychology & Major depressive disorder \\
Depression (mood) & Bipolar disorder & Schizophrenia & Anxiety \\
Anxiety disorder & Personality disorder & DSM-5 & Diagnostic and Statistical Manual of Mental Disorders \\
Classification of mental disorders & Causes of mental disorders & Antidepressant & Psychotherapy \\
Cognitive behavioral therapy & Psychosis & Post-traumatic stress disorder & Eating disorder \\
Dysthymia & Panic attack & Suicide & Social psychology \\
Personality & Generalized anxiety disorder & Borderline personality disorder & Stress \\
Mood swing & Cognitive psychology & & \\
\bottomrule
\end{tabularx}
\end{table*}
\begin{table*}[h!]
\centering
\caption{Extended set of secondary mental health topics}\label{tab:sec}
\begin{tabularx}{\textwidth}{>{\raggedright\arraybackslash}X >{\raggedright\arraybackslash}X >{\raggedright\arraybackslash}X >{\raggedright\arraybackslash}X >{\raggedright\arraybackslash}X}
\toprule
Psychiatric hospital & Mental distress & Mental toughness & Mental state & Philosophy of mind \\
Mental chronometry & Mental mapping & Mental Health Act & Orientation (mental) & Insanity defense \\
Mental age & Mental disability & Mental therapy & Mini–mental state examination & Positive mental attitude \\
Mental health inequality & Insanity & Creativity and mental health & Mental Hygiene & Mental lexicon \\
Mental reservation & Mental health triage & Mental illness in media & Mental health service & Mental health nursing \\
Narcissistic personality disorder & Mental status examination & Menstruation and mental health & Mental environment & Mental operations \\
Abortion and mental health & Mental health law & Rethink Mental Illness & Postpartum depression & Telephone phobia \\
Anxiolytic & Social anxiety & Castration anxiety & Death anxiety & Hypochondriasis \\
Social anxiety disorder & Anxiety dream & Stimulant psychosis & Substance-induced psychosis & Postpartum psychosis \\
Tardive psychosis & Caffeine-induced psychosis & Schizoaffective disorder & Folie à deux & Unitary psychosis \\
Antipsychotic & Brief psychotic disorder & List of antidepressants & Antidepressant discontinuation syndrome & SSRI \\
TCA & TeCA & SNRI & Atypical antidepressant & Pharmacology of antidepressants \\
Antidepressants and suicide risk & Hydrazine (antidepressant) & Second-gen antidepressant & Trazodone & Countries by antidepressant use \\
Fluoxetine & Tachyphylaxis & Mirtazapine & Antidepressants in Japan & Bupropion \\
Gestalt psychology & Analytical psychology & Shadow (psychology) & Educational psychology & Evolutionary psychology \\
Positive psychology & Filipino psychology & Association (psychology) & Developmental psychology & International psychology \\
Physiological psychology & Manipulation (psychology) & Doctor of Psychology & Narrative psychology & Transpersonal psychology \\
Individual psychology & Psychopathology & Biological psychopathology & Developmental psychopathology & HiTOP \\
Child psychopathology & RCAP & Development and Psychopathology & Avolition & Diathesis–stress model \\
MMPI & PANSS & Stress (biology) & Stressor & Psychological stress \\
Stress management & Stress hormone & Chronic stress & Anorexia nervosa & Anorexia (symptom) \\
Anorexia mirabilis & Anorexia athletica & Pro-ana & Sexual anorexia & Atypical anorexia nervosa \\
History of anorexia nervosa & People with anorexia & Deaths from anorexia & Cachexia & Infection-induced anorexia \\
Bulimia nervosa & Appetite & Disorganized schizophrenia & Sluggish schizophrenia & Childhood schizophrenia \\
Risk factors of schizophrenia & Evolution of schizophrenia & Religion and schizophrenia & Origin of influencing machine & People with schizophrenia \\
Anhedonia & Thought disorder & History of schizophrenia & Bipolar I disorder & Bipolar II disorder \\
People with bipolar disorder & Cyclothymia & Mood stabilizer & Lamotrigine & Sleep in bipolar disorder \\
Mood disorder & Epigenetics of bipolar disorder & Mood (psychology) & Mood congruence & Mood tracking \\
Euphoria & Mood management theory & Psychopathy & Big Five traits & Personality psychology \\
Antisocial personality disorder & Dissociative identity disorder & Schizoid personality disorder & Personality change & Enneagram of Personality \\
OCPD & Avoidant personality disorder & Histrionic personality disorder & & \\
\bottomrule
\end{tabularx}
\end{table*}

\begin{table*}[h!]
\centering
\caption{List of relation types used to represent semantic links between mental health concepts} \label{tab:normrels}
\renewcommand{\arraystretch}{1.2}
\begin{tabularx}{\textwidth}{>{\raggedright\arraybackslash}X >{\raggedright\arraybackslash}X >{\raggedright\arraybackslash}X >{\raggedright\arraybackslash}X >{\raggedright\arraybackslash}X}
\toprule
source\_or\_authority & diagnostic\_indication & clinical\_recommendation & categorization & context\_dependent \\
background\_factors & instance\_of & opposite\_of & aims\_to\_help & family\_history \\
helps\_manage & unavailable & negation & engagement & potential\_cause \\
used\_in & helps\_with\_process & occurs\_in\_context & similar\_to & prevalence \\
has\_condition & structural\_attribute & overlaps\_with & helps\_develop & comparative\_benefit \\
targets\_patients & aims\_to\_improve & treatment\_practice & treatment\_line & exclusion \\
remission\_timeline & resolves\_with & symptom similarity & part\_of & potential\_effectiveness \\
frequently\_co\_occurs & causes & illustrates & treatment\_preference & side\_effect \\
examplesinclude & onset\_age & has\_heritability & addresses & treatment\_magnitude \\
temporal\_property & defines & complicated\_by & measured\_by & comparative\_magnitude \\
created\_by & limitation & definedas & alternative\_to & likelihood \\
enhanced\_effectiveness & not only & diagnosis\_related & helps\_prevent & communication \\
adverse\_effect\_or\_outcome & has\_severity & inhibits & modifies & modulates \\
response\_rate & frequency\_pattern & aims\_to\_resolve & usage & field\_of\_work \\
requires & susceptibility & helps\_distinguish & onset & varies \\
requires\_exposure & assists\_with & recovery\_rate & treatment\_access & prescriptive\_authority \\
found\_in & has\_goal & classification & has\_model & redefined\_as \\
helps\_identify & shares\_mechanism\_with & reduced\_by & helps\_provide & has\_role \\
evidence\_based & is\_reflection\_of & historical\_finding & compared\_with & different\_from \\
has\_mechanism & complementary\_to & clinical\_expectation & theoretical\_framework & topic\_related \\
genetic\_association & synonym & induces & has\_method & aims\_to\_change \\
term\_equivalence & potential\_history & clinical\_practice & indicated\_for & belief\_claim \\
helps\_treat & occurs\_after & affected\_group & research\_focus & caused\_by \\
differential\_diagnosis & educates & conceptual\_model & eliminates & potential\_harm \\
negatively\_associated\_with & explains & has\_attribute & evidence\_detail & contraindication \\
evidence\_claim & potential\_involvement & suggests & occurs\_during & has\_part \\
aims\_to\_reduce & has\_characteristic & historical\_classification & affects & developed\_from \\
disrupts & manages & common\_pattern & prevents & occurs\_independently \\
avoidance\_level & effective & combination\_therapy & correlates\_with & exacerbates \\
authorization & has\_component & frequency & occurs\_on\_top\_of & trained\_in \\
example & treats & involved\_in & occurs\_before & subtype\_of \\
key\_factor\_for & related\_to & early\_onset & has\_type & is\_a \\
needs\_adaptation & has\_diagnosis & not\_caused\_by & common\_occurrence & used\_for \\
cannot\_occur\_during & origin & follows & helps\_change & represents \\
potential\_benefit & avoids & current\_status & outcomes & untreated\_outcome \\
shows & delivery\_method & & & \\
\bottomrule
\end{tabularx}
\end{table*}

\end{document}